\documentclass[10pt,twocolumn,letterpaper]{article}
\pdfoutput=1
\usepackage{iccv}
\usepackage{times}
\usepackage{epsfig}
\usepackage{graphicx}
\usepackage{amsmath}
\usepackage{amssymb}
\usepackage{enumitem}
\usepackage{balance}

\iccvfinalcopy 

\def\iccvPaperID{1855} 

\ificcvfinal\fi
\begin{document}

\title{Unconstrained Foreground Object Search}

\author{Yinan Zhao$^*$,
Brian Price$^+$,
Scott Cohen$^+$,
Danna Gurari$^*$\\
{\small $~^*$ University of Texas at Austin,} 
{\small $~^+$ Adobe Research}\\
{\tt\small yinanzhao@utexas.edu, \{bprice,scohen\}@adobe.com, danna.gurari@ischool.utexas.edu} 
}

\maketitle
\thispagestyle{empty}

\begin{abstract}
Many people search for foreground objects to use when editing images.  While existing methods can retrieve candidates to aid in this, they are constrained to returning objects that belong to a pre-specified semantic class.  We instead propose a novel problem of unconstrained foreground object (UFO) search and introduce a solution that supports efficient search by encoding the background image in the same latent space as the candidate foreground objects.  A key contribution of our work is a cost-free, scalable approach for creating a large-scale training dataset with a variety of foreground objects of differing semantic categories per image location.  Quantitative and human-perception experiments with two diverse datasets demonstrate the advantage of our UFO search solution over related baselines.
\end{abstract}

\section{Introduction}
Image-based search, the task of retrieving images based on an image query, is a popular research problem with many applications~\cite{rodriguez2016data,torralba200880,ak2018learning,yang2014context,hays2007scene}.  While it is often used to find visually or semantically similar images to the query image, a less explored subproblem in this domain is searching for content to edit the query image.  Yet the importance of this subproblem is evidenced by the existence of many stock image websites, for example \texttt{shutterstock.com}, \texttt{www.istockphoto.com}, and \texttt{stock.adobe.com} to name a few, which contain tens of millions of images of objects on a white or plain background to make it easy to cut out just the foreground object to use it in another image.  Whether a user is placing an object on top of a complete image (compositing) or using an object to partially fill a hole (created, for example, by removing another object or area), an important part of the creative process is to find a large variety of content that is compatible with the surrounding background in order to explore multiple possible outcomes.

The most relevant related work to this subproblem are compositing-aware methods which require a user to specify the desired object type to be pasted into a query image, and then search for suitable objects~\cite{lalonde2007photo,zhao2018compositing}.\footnote{Of note, search is a valuable approach since deep learning based methods that synthesize realistic-looking content are unable to do so for large holes with complex surrounding structures~\cite{pathak2016context,yang2016high,iizuka2017globally,yu2018generative}.}   While specifying the object type to be inserted can guide the search process, it also introduces a limitation that creatives cannot explore many possible image modifications representing a variety of objects that can be inserted into a query image believably.

\begin{figure}[t!]
\centering
\includegraphics[width=0.49\textwidth]{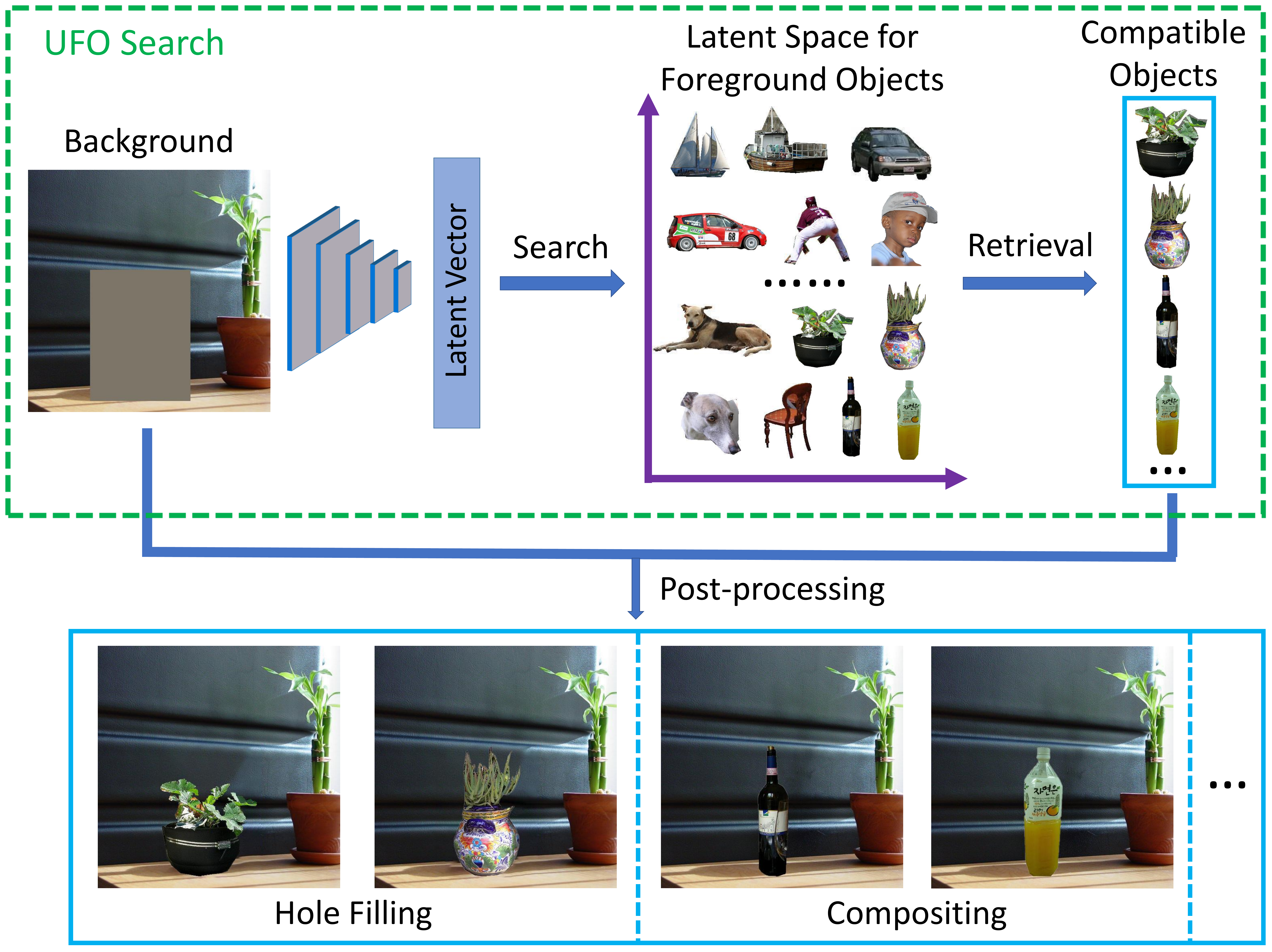}
\caption{We propose a method to search for foreground objects that are semantically compatible with a background image. In this example, our approach takes the background image with a hole on the table, searches in a large object database consisting of multiple semantic classes, and returns compatible foreground objects.  This example illustrates how UFO search can be used for hole filling (using~\cite{Barnes:2009:PAR} to fill in the gaps around the object) and compositing.
}
\label{fig:teaser}
\end{figure}

In this paper, we propose the problem of unconstrained foreground object search (UFO search).  Specifically, the goal is to \emph{search for foreground objects that are semantically compatible with a background image} without any constraint on what objects to retrieve.  
An object is compatible with a background image if it can be realistically composited into the image or used to aid hole filling, as illustrated in Figure~\ref{fig:teaser}.  Here, we focus on \emph{semantic} compatibility as other methods address correcting geometrical errors~\cite{lin2018st,azadi2018compositional} and low-level color and appearance differences~\cite{tsai2017deep,zhu2015learning}.

We also introduce a novel solution for UFO search.  Inspired by~\cite{zhao2018compositing}, our network projects background images and foreground objects into a high-level feature space, \emph{without} requiring object labels, such that compatible objects and backgrounds are near each other.  These high-level features are then used for efficient search.  A key contribution of our work is a cost free, scalable approach for creating a large (noisy) dataset for training unconstrained foreground object search methods.  Experiments demonstrate the effectiveness of our UFO search method over numerous related baselines.

\section{Related Work}
\noindent
\textbf{Constrained Foreground Object Search} is the task of retrieving foreground objects that are compatible with the background image given the desired object type. Early works such as Photo Clip Art \cite{lalonde2007photo} retrieved foreground objects of a given class based on handcrafted features such as camera orientation, lighting, resolution and local context. More recently, Tan et al. \cite{tan2018and} used off-the-shelf deep CNN features from the context to find suitable foreground persons particularly for person composition. Zhao et al. \cite{zhao2018compositing} used end-to-end feature learning to adapt to different object categories.
In contrast, our approach has no constraint on what objects to retrieve and our experiments demonstrate it can retrieve compatible object candidates of different classes.
\vspace{0.2em}

\noindent
\textbf{Predicting Compatibility.}
Prior work~\cite{zhu2015learning} has demonstrated it is possible to solve a related problem of predicting whether a composite and image are compatible.  However, while \cite{zhu2015learning} focuses on low-level compatibility (e.g., color, lighting, texture), we aim to stay largely agnostic to low-level properties (since properties such as lighting and color differences can be corrected in post-processing) and instead address semantic compatibility.  Experiments show the advantage of our solution over \cite{zhu2015learning} for the UFO search task.

\vspace{0.2em}
\noindent
\textbf{Context-based Reasoning} has been used in object recognition and detection \cite{divvala2009empirical}.  Some works model the interaction of existing content in the image.  For example, early works \cite{bar1996spatial,strat1991context} incorporated context cues for object recognition and Bell et al. \cite{bell2016inside} recently proposed a recurrent neural network for object detection.  Our method more closely aligns with methods that make predictions about missing content based on image context.  For example, one work proposes solving object detection based on context cues only~\cite{torralba2003contextual}.  Another work trains a standalone object-centric context representation to detect missing objects~\cite{sun2017seeing}.  While these methods focus on the binary decision of whether there should be an object of a semantic class at a specific location, our approach addresses a distinct problem of searching for foreground object instances that are compatible with the context.  Moreover, the compatible foreground objects may be a subset of a semantic class or come from different classes. 
\vspace{0.2em}

\noindent
\textbf{Scene Completion methods~\cite{hays2007scene, IntRetrieval2009, zhu2015faithful}}, like our work, involve inserting foreign content into an image.  However, such methods address a distinct problem from our proposed UFO search problem.  The former assumes the goal is to find a \emph{patch} to insert into a scene image.  Consequently, it must find a patch that seamlessly matches every background element in the scene.  In contrast, UFO Search only finds a \emph{compatible object}.  This distinction provides an advantage over Scene Completion methods since UFO search methods can work in a general-purpose pipeline that positions a foreground object over the majority of the hole, and then applies any downstream post-processing methods (exemplified in Figure~\ref{fig:teaser}) to fill the gaps.  
\section{Methods}
\begin{figure*}[!ht]
\centering
\includegraphics[width=1.0\textwidth]{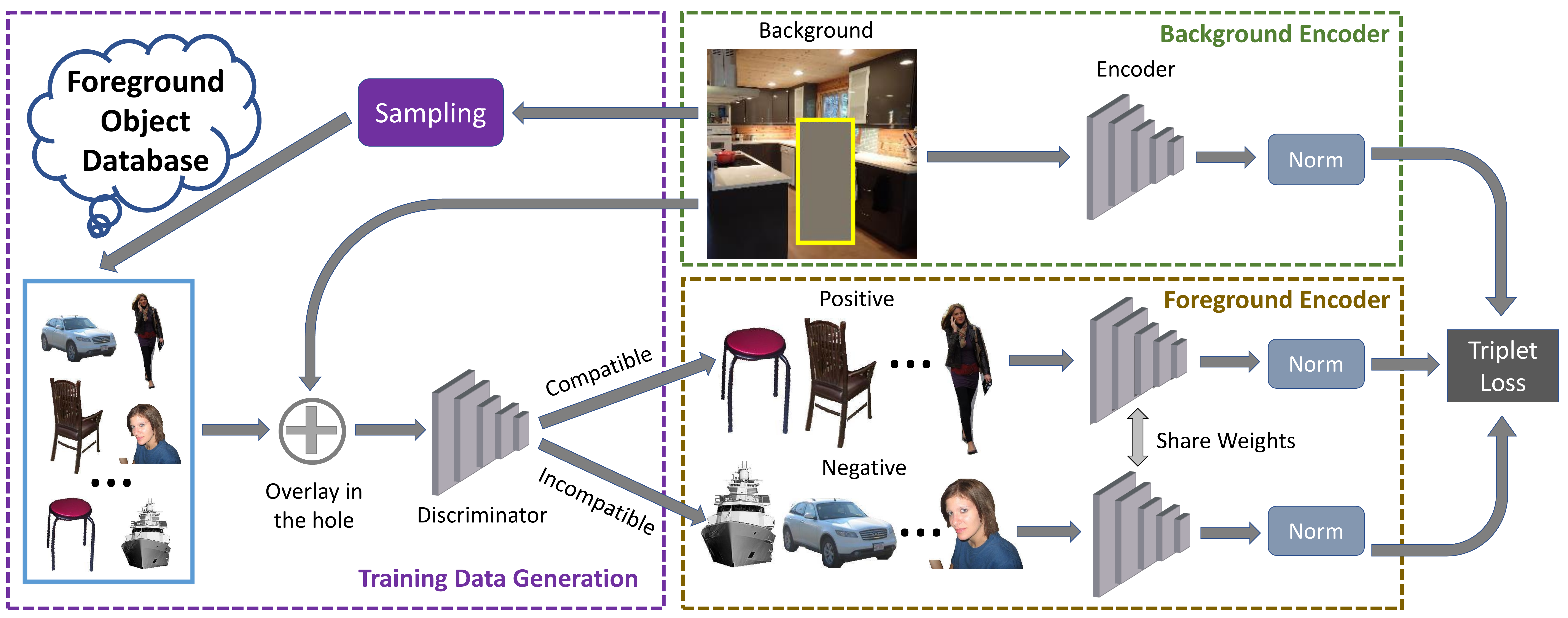}
\vspace{-1.5em}
\caption{Architecture and training scheme of UFO search. Given a background image with a hole, we first sample foreground objects to overlay in the hole. Then the pretrained discriminator takes the overlaid image and identifies compatible and incompatible foreground objects. We use two encoders to encode the background image and foreground objects respectively. The triplet loss encourages the compatibility between the background and positive samples to be larger than its compatibility with negative samples.}
\label{fig:method_overall}
\end{figure*}

We propose a method for retrieving foreground objects from a database that are semantically compatible with a given image at a specified location.  Our approach learns how to represent both the background image and each candidate foreground object in a shared search space that supports efficiently ranking the compatibility of all foreground objects.  The architecture and training scheme for our approach are summarized in Figure~\ref{fig:method_overall} and described below.  

\subsection{Deep Learning Architecture}\label{method:arch}
We propose a deep neural network that consists of two encoders which characterize the background image and foreground objects respectively by projecting them into a high-level feature space where compatible objects and image are near each other spatially.  The approach is inspired by~\cite{zhao2018compositing}, though our architecture is more straight-forward and does not require an object label. The input to the foreground encoder is a foreground object on the background of mean image value, and the input to the background encoder is the background image with a hole\footnote{The hole is filled with the mean image value.}  (needed for masking out the original object at that location in the training set) at the desired object location. The high-level feature outputs from the foreground objects can be stored in an index so that the objects can be retrieved given the feature corresponding to a background image.  

Both encoders are derived from the popular VGG-19~\cite{simonyan2014very} architecture (up to \textit{fc6} layer), that takes as input images of size $224 \times 224$ and outputs 4096 dimensional feature embeddings.  For the foreground object encoder, our goal is to capture the semantics of foreground objects.  Since that is already captured well in the VGG-19~\cite{simonyan2014very} architecture, we keep the weights that were pretrained for the ILSVRC-2014 competition~\cite{russakovsky2015imagenet} fixed during training.  In contrast, for the background encoder, we initialize the weights with those pretrained for the ILSVRC-2014 competition~\cite{russakovsky2015imagenet} and then modify them during training.  The encodings of the background image and foreground objects are then converted to unit feature vectors with $\ell_2$ normalization and used to compute compatibility, by measuring their cosine similarity.

\subsection{Loss Function}\label{method:loss}
We adopt as our loss function a triplet loss~\cite{wang2015unsupervised} that takes as input a background image, positive sample, and negative sample.  This function encourages the compatibility between a background image and a good foreground object (i.e., positive sample) to be larger than its compatibility with a bad foreground object (i.e., negative sample).  Formally, given a background image $I_b$, positive sample $I_f^p$, and negative sample $I_f^n$, we want to enforce $C(I_b,I_f^p)>C(I_b,I_f^n)$. 
The triplet loss is a hinge loss
$L(I_b,I_f^p,I_f^n)=\max(0,  C(I_b,I_f^n)+M-C(I_b,I_f^p))$
where $M$ is a positive margin to encourage a gap between the positive and negative sample.  The training objective is to minimize the loss over all the sampled triplets. 

\subsection{Training Data Generation} \label{method:trainingDataGeneration}
We generate a training dataset that consists of triplets that contain a (1) background image, (2) compatible foreground object (positive), and (3) incompatible foreground object (negative).  Exemplar triplets are shown in Figure~\ref{fig:method_overall}.  

Our key challenge lies in how to generate a sufficient number of positive samples per background image.  That is because, for each background image, we only have one known positive sample: the foreground object that originally was there.  Yet, for many scenes, numerous other foreground objects are plausible.  We introduce two mechanisms for identifying a diversity of compatible foreground objects per background image: a discriminator to identify a noisy set of compatible foreground objects for each background image and a sampling module to accelerate identifying plausible foreground objects for training the encoder.

\vspace{-1em}
\paragraph{Training Data Filtering.}
We propose a discriminator to help filter the training data for effective training samples.  We design it to take as input a given background image with the foreground object overlaid in the hole and output a prediction of whether they are compatible.  Note that this discriminator is distinct from that employed for our UFO search encoder (described in Section~\ref{method:arch}).  While our UFO search encoder learns how to represent the foreground objects and background image \emph{de-coupled} in a complex, high-level feature space, the discriminator instead takes them \emph{coupled} as input, with the foreground object overlaid on the background image.  Consequently, while our UFO search encoder returns an efficient representation for search where objects that are compatible are close and objects that are not compatible are far away, the discriminator outputs a \textquotedblleft yes\textquotedblright or \textquotedblleft no\textquotedblright answer for a single pair of a foreground object and background image.  We will show in Section~\ref{sec:evaluation} that the discriminator alone is unsuitable for solving our compatibility problem (in terms of accuracy and speed) but is valuable for boosting the performance of our UFO search encoder by generating noisy yet richer training triplets.

\begin{figure}[!t]
\centering
\includegraphics[width=0.49\textwidth]{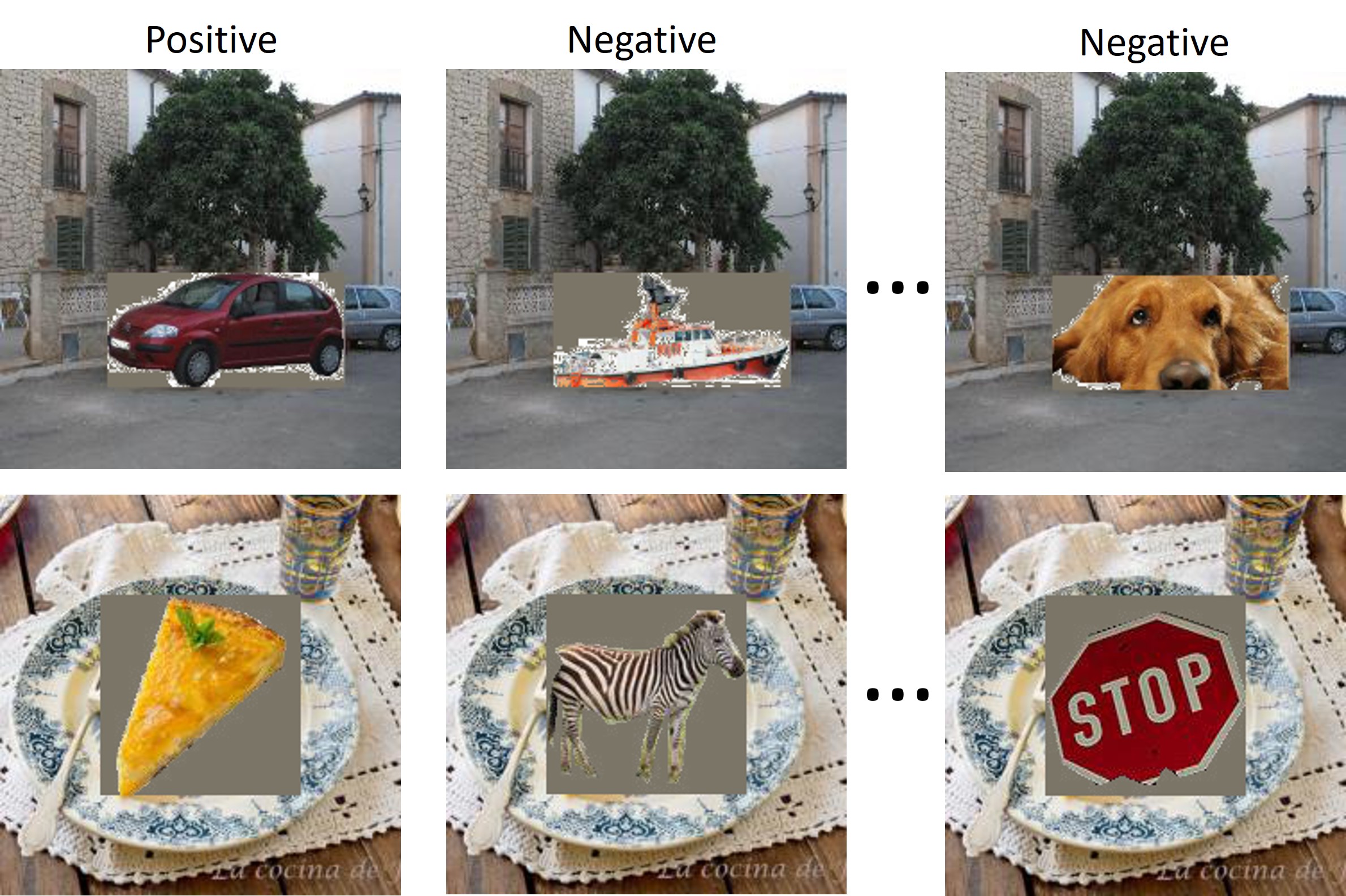}
\caption{Examples of positive and negative samples used to train the discriminator for compatibility prediction. The positive samples (the left column) are created by overlaying the original object in the hole. The foreground objects in negative samples (the middle and right column) are randomly sampled from other images. }
\label{fig:D_input}
\end{figure}

For the discriminator's architecture, we adapt VGG-19~\cite{simonyan2014very} by replacing the last fully connected layer to produce a scalar value that indicates the compatibility score.  To encourage the network to utilize high-level features so it focuses on semantic compatibility, we initialize with the weights pretrained for the ILSVRC-2014 competition~\cite{russakovsky2015imagenet} and freeze all the convolutional layers.  We train all the fully connected layers from scratch using a sigmoid cross-entropy loss.  For training data, we generate compatible training examples by overlaying the original foreground object in the hole, and generate incompatible examples by selecting a random foreground object from another background image, resizing the object to fit in the hole, and then positioning it at the center of the hole.  Examples of compatible and incompatible samples that we feed to train our discriminator are shown in Figure~\ref{fig:D_input}. Note that in the hole we overlay the object alone rather than the original patch containing the object. Otherwise the discriminator will simply learn to use low-level cues such as boundary continuity rather than semantics for classification.

We restrict training triplets to only include foreground objects that the discriminator confidently deems are (in)compatible when training the encoder.  A foreground is deemed compatible with a given background if the discriminator predicts the compatibility score to be higher than a threshold $t_{high}$ and incompatible if the score is lower than a threshold $t_{low}$.  Despite training with a single ground truth object per background image, we show in the experiments that the discriminator can sufficiently rank the compatibility of diverse foreground objects.  The success of training a classifier to rank has similarly been observed in prior work, e.g. Zhu. et.al~\cite{zhu2015learning} for the task of ranking the realism of image composites by low-level appearance.

\vspace{-1em}
\paragraph{Collecting Candidate Positive Examples Faster.} 
While the discriminator solves an easier task than our UFO search method by solving a \textquotedblleft yes" or \textquotedblleft no" problem for a coupled input, it does so at the expense of efficiency.  That is because naively applying the pretrained discriminator can require comparing each background image against almost every foreground object in a database before locating a sufficient number of high scoring compatible examples.

To speed up the discriminator's role in generating training data, we introduce two heuristics for sampling plausible foreground objects.  First, we retrieve the top $K_C$ most similar background scenes, and put the objects within those scenes into the sample set.  The assumption is that similar backgrounds are likely to offer (possibly a diversity of) compatible objects.  For example, for a given grass scene, we can find similar scenes such as a picnic on a lawn.  The sitting persons or folding chairs in the picnic scene are also likely to be compatible with the grass scene.  Second, we sample the top $K_G$ foreground objects that are most similar to the original object, motivated by the assumption they are more likely to be compatible with the given context.  For example, if a dog is running on the grass in the original image, it is likely that dogs in other scenes will also be compatible.  In a database of over 60,000 objects, we observe a more than 20x speed up from the two proposed heuristics (from 731 to 32 random samples on average) to find another compatible object other than the original object in the hole.

\subsection{Implementation}
At training time, we employ the Adam solver~\cite{kingma2014adam} with fixed parameters $\beta_1=0.5$ and $\beta_2=0.999$.  The initial learning rate is set to $l_r=0.00001$ to train the encoder and $l_r=0.00002$ to train the discriminator.  We set the positive margin $M$, which encourages a gap between the positive and negative sample, to 0.3, with the threshold $t_{high}$ for identifying positive samples in compatibility prediction set to 0.8 and the threshold $t_{low}$ for identifying negative samples in compatibility prediction set to 0.3.  All the background and foreground input images are set to a size of $224\times224$.  We train the discriminator beforehand and then fix the discriminator when training the encoder.  Training with PyTorch~\cite{paszke2017automatic} takes 63 hours for 142,300 iterations on a single NVIDIA GeForce GTX 1080 Ti card.  

At test time, we apply the background encoder to retrieve the most compatible foreground objects for a given background image with a hole. Compatible objects are found using nearest neighbor search between features describing the background image and foreground object.  We speed up nearest neighbor search by using Faiss~\cite{JDH17} to build an index for the evaluation set of foreground objects. After the speedup, it takes \textless0.1 seconds to retrieve top 25 compatible objects from a database of over 10,000 objects.
\section{Experiments}\label{sec:evaluation}
We now examine the power of our UFO search approach in finding compatible foreground objects for a given hole in a background image.  We examine the following questions: (1) How often do related baselines re-purposed for UFO search retrieve compatible foreground objects?, (2) How often does our UFO search method retrieve compatible foreground objects?, and (3) How do our different design decisions contribute to the performance of our method?  We conduct experiments on two datasets, with a quantitative evaluation in Section~\ref{exp_cais} and user evaluation in Section~\ref{exp_coco}. 

\subsection{Quantitative Evaluation}\label{exp_cais}
We first conduct experiments using background images with holes that have various positions and sizes.

\vspace{-1.35em}
\paragraph{Dataset:}  
Since large-scale datasets identifying all compatible foreground objects for a background image with a hole do not exist, we use as a proxy the image compositing dataset CAIS~\cite{zhao2018compositing}.  CAIS contains background images with a hole, an assigned category for the type of object that should fill the hole, and at least one compatible foreground object in that category.  Although designed for constrained foreground object search, CAIS is also valuable for the more general problem of UFO search since most background images with holes unambiguously match only one object category from the eight foreground object categories represented\footnote{boat, bottle, car, chair, dog, wall painting, person, and plant}.  The training set contains one compatible object for each of the 86,800 background images, using the original object in each image.  The test set contains $\sim$16-140 compatible objects per image for 80 background images, with 10 background images for each object category.  

\begin{table*}[!ht]
\begin{center}
\begin{tabular}{l | c c c c c c c c | c}
\hline
Method & boat & bottle & car & chair & dog & painting & person & plant & overall \\
\hline\hline
Shape~\cite{zhao2018compositing} & 7.47 & 1.16 & 10.40 & 12.25 & 12.22 & 3.89 & 6.37 & 8.82 & 7.82\\
RealismCNN~\cite{zhu2015learning} & 12.33 & 7.19 & 7.55 & 1.81 & 7.58 & 6.45 & 1.47 & 12.74 & 7.14\\
CFO-C Search~\cite{zhao2018compositing} & 57.48 & 14.24 & 18.85 & 21.61 & 38.01 & \textbf{27.72} & \textbf{47.33} & 20.20 & 30.68\\
CFO-D Search~\cite{zhao2018compositing} & 55.48 & 8.93 & 24.10 & 18.16 & \textbf{57.82} & 21.59 & 27.66 & 23.13 & 29.61\\
Ours: UFO Search  & \textbf{59.73} & \textbf{21.12} & \textbf{36.63} & 19.27 & 36.51 & 25.84 & 27.11 & \textbf{31.19} & \textbf{32.17} \\
\hline
Ours: No BG Training  & 49.09 & 0.62 & 3.23 & 9.01 & 7.37 & 11.66 & 7.30 & 22.02 & 13.79 \\
Ours: No Discriminator & 58.07 & 17.22 & 20.71 & \textbf{21.93} & 37.05 & 24.57 & 27.11 & 25.05 & 28.97 \\
Ours: Discriminator Only & 48.71 & 8.35 & 21.42 & 17.32 & 50.61 & 20.28 & 22.14 & 17.35 & 25.77 \\
Ours: Regression  & 55.33 & 9.90 & 18.31 & 17.42 & 27.79 & 23.76 & 35.66 & 10.83 & 24.87 \\
\hline
\end{tabular}
\end{center}
\vspace{-0.6em}
\caption{\textit{Mean Average Precision} for the top 100 retrievals of four baselines, our UFO search method, and its four ablated variants.}
\label{table:MAP1}
\end{table*}

\vspace{-1.35em}
\paragraph{Baselines: } We compare our approach to four baselines:

\textit{Shape}~\cite{zhao2018compositing}: This adopts a naive strategy of ranking compatibility based on the extent to which the foreground object's aspect ratio (i.e., width/height) matches the hole's aspect ratio.  For example, a tall hole would match a tree or pedestrian better than a car.  

\textit{RealismCNN}~\cite{zhu2015learning}: It uses a discriminator to predict the realism of image composites in terms of low-level cues such as color, lighting, and texture compatibility.  After overlaying each foreground object into the hole (as in Figure~\ref{fig:D_input}), the pretrained model ranks the compatibility of all objects.

\textit{Two constrained search methods}: Since \textit{constrained} search methods require a category as input and so are not directly useful, we examine two ways to adapt them for an \textit{unconstrained} setting.  First, we train a classifier to decide which category to fill in the hole\footnote{The classifier employs the VGG architecture with weights pretrained on ImageNet, and achieves overall accuracy of 63.75\%.} and then apply a constrained search method to retrieve suitable instances within that category.  We call this \textit{Constrained Foreground Object Search - Classifier (CFO-C Search)}.  Note that it has the limitation that it requires collecting class labels to train the classifier and so would not recover from the errors of the classifier.  The second approach retrieves the top 100 objects for each of the eight categories using the constrained search method and then applies our trained discriminator to rank the retrieved 800 (100x8) objects.  We call this \textit{Constrained Foreground Object Search - Discriminator (CFO-D Search)}.  Note that \textit{CFO-D Search} becomes less practical with more categories and more retrievals, because it requires expensively traversing every retrieval with the discriminator and ranking all the retrievals.  We evaluate both approaches using the constrained search algorithm \cite{zhao2018compositing}.

\vspace{-1.35em}
\paragraph{Ablated Variants:} We evaluate ablations of our \textit{UFO Search} to assess the influence of different design decisions:
\vspace{-1.5em}
\begin{description}[style=unboxed,leftmargin=0cm]
\setlength\itemsep{0em}
\vspace{-0.25em}
\item - \textit{No BG Training}: It uses the pretrained weights for the ILSVRC-2014 competition~\cite{russakovsky2015imagenet} as the background encoder's weights.  This is valuable for assessing the benefit of training the background encoder when training UFO search.
\vspace{-1.5em}
\item - \textit{No Discriminator}: It does not use our training data generation scheme, described in Section~\ref{method:trainingDataGeneration}.  Instead, it uses one compatible foreground object per background image (i.e., the original one in the hole) and many incompatible samples (i.e., all foreground objects in other background images).  
\vspace{-0.4em}
\item - \textit{Discriminator Only}:  The discriminator described in Section~\ref{method:trainingDataGeneration}, which we use for training data generation, is instead used to predict compatibility at test time.  Recall that a limitation of this approach is that it requires overlaying each foreground object in the hole of each test background image, which is very computationally expensive.
\vspace{-0.4em}
\item - \textit{Regression}: This approach matches the \textit{No Discriminator} approach except that it trains for the regression problem (i.e., using \textit{Mean Square Error (MSE)}) instead of the ranking problem (i.e., using the triplet loss).  To do so, it regresses to the feature of the original foreground object in the hole from the background image using the MSE loss function.  We evaluate on a simplified situation (without the discriminator) to assess the training approach on its own.
\end{description}

\vspace{-2em}
\paragraph{Evaluation Metrics: } 
We use \textit{mean Average Precision} (mAP) for evaluation, which is a common metric in image retrieval. We report mAP for each category as well as overall, by averaging over all category mAPs.  To make our findings compatible with the constrained foreground object search methods (\textit{CFO-C Search} and \textit{CFO-D Search}), we evaluate the mAP for the top 100 retrievals.  This is because CFO methods do not rank all objects in all categories.  We share the mAP results with respect to all the retrievals for all other methods in the Supplementary Materials.

\vspace{-1.35em}
\paragraph{Overall Results: }
Results are shown in Table \ref{table:MAP1}.  

Overall, our \emph{UFO Search} method outperforms the four related baselines: \emph{Shape}, \emph{RealismCNN}~\cite{zhu2015learning}, \emph{CFO-C}, and \emph{CFO-C}.  For example, mAP is \textbf{32.17\%} for \emph{UFO Search}, which is over 24 percentage points better than for \emph{Shape} and \emph{RealismCNN}.  These results reveal that relying on hole shape alone or low-level compatibility alone is not very informative, and demonstrates the advantage of addressing semantic compatibility directly.  \emph{UFO Search} also results in a 1.49 percentage point improvement over the next best constrained search baseline.  This shows that \textit{UFO Search} not only offers a scalable end-to-end solution that avoids requiring a separate class predictor (required by \emph{CFO-C}) or large computational costs (required by \emph{CFO-D} as more categories and retrievals are considered), but also yields improved prediction accuracy.  This highlights a benefit of directly learning to solve the \textit{unconstrained} search problem rather than modifying \textit{constrained} search methods.   

Our analysis also shows how our \textit{UFO Search} compares to the baselines for different object categories.  As shown in Table \ref{table:MAP1}, our \textit{UFO Search} outperforms all baselines on the following four object categories: boat, bottle, car, and plant.  The top-performer for the other four categories is shared between three baselines.  One reason our \textit{UFO Search} performs poorer at times is that for the \textit{person} category it can mistakenly retrieve boats for surfing scenes and dogs for park scenes.  Our \textit{UFO Search} also at times mistakenly retrieve boats and chairs for the \textit{painting} category.  Additionally, for the \textit{dog} category, it at times mistakenly retrieve cars and persons for street scenes.  These findings suggest that our approach understands the context semantically, but does not always capture well the potential interaction between the inserted object and the context for specific categories such as \textit{person}, \textit{painting}, and \textit{dog}.  We hypothesize discriminator based methods (\textit{CFO-D Search} and \textit{Ours: Discriminator Only}) can perform better than our \textit{UFO Search} method for the \textit{dog} category because it can be easier to recognize which foreground objects are incompatible for the hole's size and shape when overlaying the foreground object directly in the hole (as the discriminator methods do).

\vspace{-1.35em}
\paragraph{UFO Design Analysis Results: }
Results in Table~\ref{table:MAP1} also illustrate the benefit of design choices for our \emph{UFO Search}.

\begin{figure}[!th]
\centering
\vspace{-0.5em}
\includegraphics[width=0.48\textwidth]{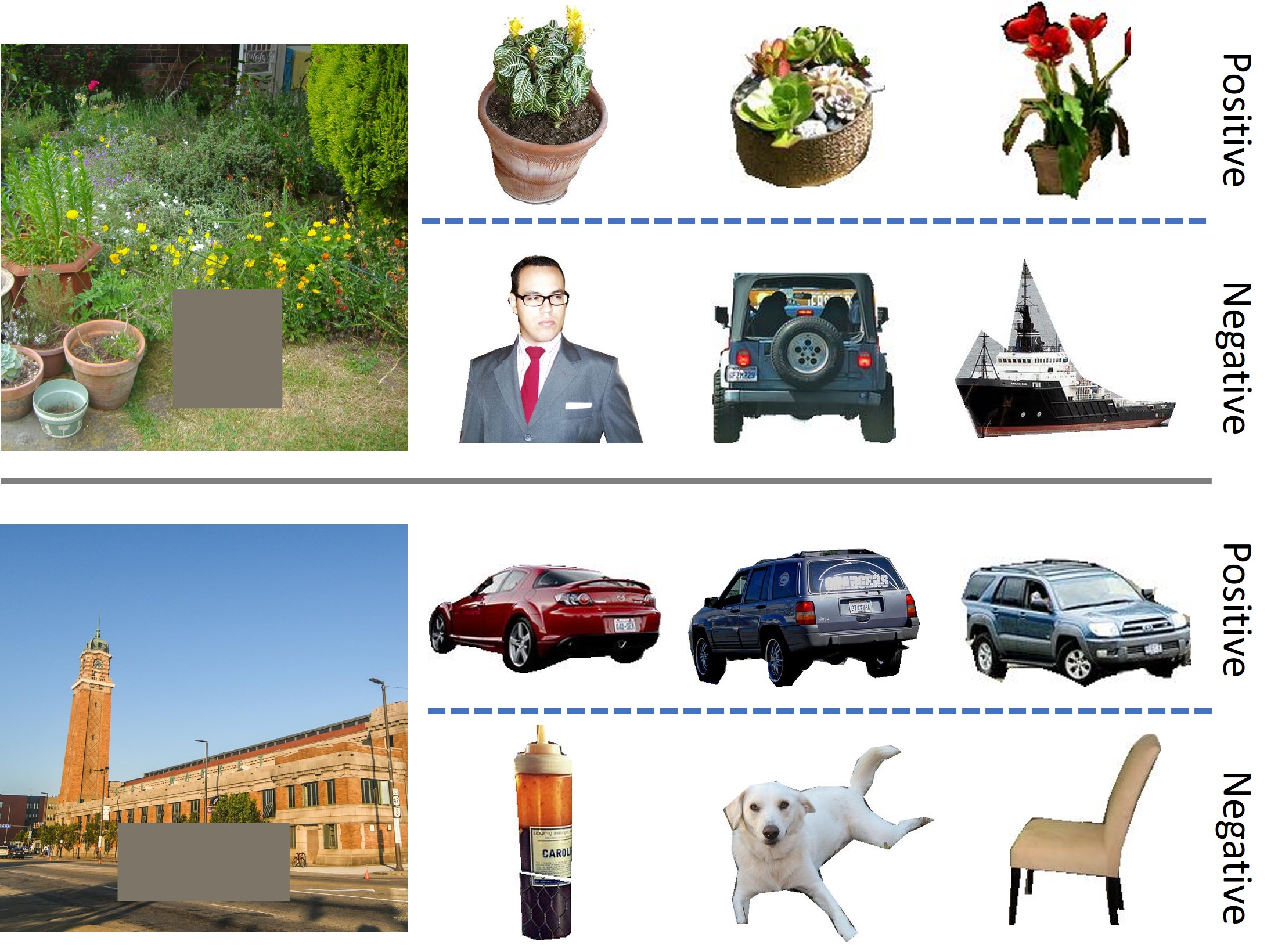}
\vspace{-1.5em}
\caption{Positive and negative samples that are deemed compatible and incompatible with the background image by our discriminator.  As shown, it can identify multiple compatible objects. }
\vspace{-1em}
\label{fig:D_pos_neg}
\end{figure}

The poor performance from the \emph{Discriminator Only} demonstrates that the triplets sampled by the discriminator are imperfect; i.e., mAP score is 25.77\%.  Moreover, it performs worse than our \emph{UFO Search} both in terms of accuracy (i.e., mAP score is 6.4 percentage points worse) and speed (i.e., we observe over a 3000x slow down from 0.1 to 365.6 seconds when relying on the discriminator instead of \emph{UFO Search} to perform retrieval from a database of 60,000 objects).  These findings highlight a strong advantage of learning how to represent the background image and foreground objects de-coupled in a complex, high-level feature space that supports efficient search, as our \emph{UFO Search} does, rather than coupling the background image and foreground object as input, as the discriminator requires.

The gain of \emph{UFO Search} over \emph{No Discriminator} demonstrates the advantage of employing our training data generation method; i.e., we observe more than a three percentage point boost.  The encoder in \emph{UFO Search} benefits from learning using the noisy training data sampled by the discriminator.  Figure~\ref{fig:D_pos_neg} exemplifies that the discriminator can identify multiple compatible objects, despite having trained with a single positive ground truth per background image.  

The weaker performance of \emph{Regression} versus \emph{No Discriminator} illustrates the advantage of training for the ranking problem (by using the triplet loss); i.e., \textit{No Discriminator} yields more than a four percentage point boost over \emph{Regression}.  We attribute this performance gain to a benefit of training with both positive and negative samples in the ranking problem, rather than only positive samples when training for regression.  Training with both positive and negative samples better shapes the feature space by pushing compatible objects closer to the background encoding and incompatible objects farther away from the background encoding.  

Finally, the poor performance of \emph{No BG Training} compared to the three methods that train the background encoder (i.e., \emph{UFO Search}, \emph{No Discriminator}, \emph{Regression}) demonstrates the benefit of feature learning. 

\subsection{User Evaluation}\label{exp_coco}
We next conduct user evaluation on a more diverse dataset consisting of 79 object categories.  

\vspace{-1.35em}
\paragraph{Dataset: }
We employ MS-COCO~\cite{lin2014microsoft} to create a diverse dataset of 79 foreground object categories.  We use the object segmentation mask annotations to decompose each image into a background scene and foreground objects (see Supplementary Materials for more details).  This yields 14,350 background images and 61,069 foreground objects.  We use 14,230 background images for training and the remaining 120 for evaluation.  To provide a large foreground object database at test time, we use all 61,069 foreground objects in both training and testing.  This is acceptable since we do not learn the feature space for foreground objects.  In order to evaluate the effectiveness in encoding the context exclusively, we fix the hole size and position for all background images.  Specifically, we create the holes for each background scene by removing a square that bounds each foreground object.  Then we resize each background image to $224\times224$ with a hole of size $112\times112$ in the center.  

\vspace{-1.35em}
\paragraph{Ablated Variants:} We compare with our \textit{UFO Search} the four ablated variants described in Section~\ref{exp_cais}: \emph{No BG Training}, \emph{Regression}, \emph{No Discriminator}, and \emph{Discriminator Only}. 

\vspace{-1.35em}
\paragraph{Evaluation Metrics: } Since this dataset does not identify multiple compatible foreground objects per background image, we conduct a user study to measure the Precision@K (P@K), which is the percentage of compatible foreground objects in the top $K$ retrievals.  We show users a background image and $K$ candidate foreground objects retrieved by an image search approach. Users are asked to select the foreground objects that are not compatible with the background image. Each background image is evaluated by 3 different users. If any user labels a foreground object as incompatible, the foreground object is considered to be incompatible.

\begin{table}[!th]
\begin{center}
\begin{tabular}{c | c c c c c}
\hline
Method & P@5 & P@10 & P@15 & P@20 & P@25\\
\hline\hline
No Training & 12.67 & 13.33 & 13.28 & 12.50 & 12.50\\
Regression & 30.33 & 30.75 & 30.39 & 30.50 & 30.40 \\
No D & 38.50 & 36.58 & 36.11 & 35.54 & 35.57\\
D Only & 36.33 & 37.25 & 36.00 & 35.46 & 35.77\\
UFO Search & \textbf{41.83} & \textbf{40.33} & \textbf{39.39} & \textbf{38.96} & \textbf{38.83}\\
\hline
\end{tabular}
\end{center}
\vspace{-0.5em}
\caption{User study results showing the percentage of retrieved foreground objects in the top K retrievals that are deemed compatible by users. \emph{No D} = \emph{No Discriminator}, \emph{No Training} = \emph{No BG Training}, \emph{D Only} = \emph{Discriminator Only}.   
}
\label{table:coco}
\end{table}

\begin{figure*}[!ht]
\centering
\includegraphics[width=0.9\textwidth]{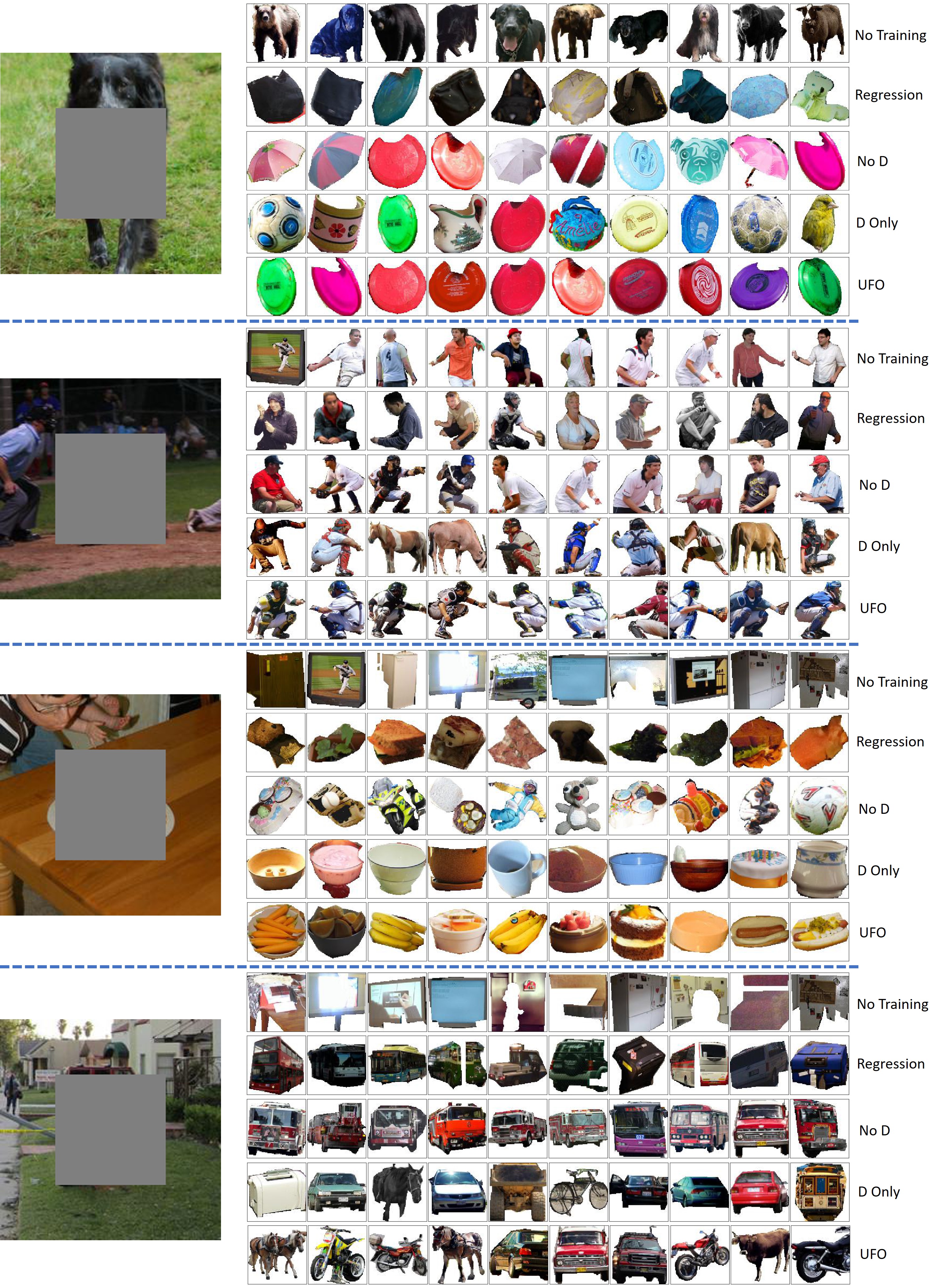}
\caption{Qualitative results on MS-COCO~\cite{lin2014microsoft}. For the top two examples, our approach retrieves objects from the only object type from MS-COCO (frisbee and catcher, respectively) that is really compatible with the context. The bottom two examples demonstrate that our approach has the potential to retrieve compatible objects of different categories when many object types are appropriate for the scene.}
\label{fig:results}
\end{figure*}

\vspace{-1.35em}
\paragraph{Overall Results: }
Quantitative results are shown in Table~\ref{table:coco} for using our \emph{UFO Search} for the top 5, 10, 15, 20, and 25 retrievals respectively.  Qualitative results are shown in Figure~\ref{fig:results}.  The top two examples illustrate that our \textit{UFO Search} can retrieve only one type of object when only one object type is compatible; specifically, it retrieves only frisbees and catchers for the dog and baseball field respectively.  Also shown is that our \textit{UFO Search} can retrieve compatible objects that are from different categories when numerous object types are appropriate for the scene.  Specifically, our approach retrieves carrots, oranges, bananas, cakes, sandwiches and hot dogs for a hole on a plate on a table (second to bottom example) and retrieve horses, motorbikes, cars and cows for the hole in the grass (bottom example).  

\vspace{-1.35em}
\paragraph{UFO Design Analysis Results:}
Our \emph{UFO Search} method outperforms all its ablated variants for every retrieval size, increasing the search precision by \textbf{3.33}, \textbf{3.08}, \textbf{3.28}, \textbf{3.42}, and \textbf{3.06} percentage points compared to the next best ablated variant in top 5, 10, 15, 20, 25 retrievals respectively.  This aligns with and reinforces our findings in Section~\ref{exp_cais}.  
    
Qualitative comparisons in Figure~\ref{fig:results} illustrate strengths of our design choices.  For the first example, while the top retrievals of \textit{UFO Search} are all compatible frisbees, the \emph{No Discriminator} retrieves umbrellas which have similar shape to frisbees but are not compatible in the context.  For the second baseball field example, all ablated variants accurately retrieve the person category, however only our \textit{UFO Search} method recognizes that a catcher is the only suitable activity for the context.  The last example shows the retrievals of \emph{Discriminator Only} can be noisy, containing incompatible objects such as a toaster, but also is effective in retrieving compatible objects of multiple categories, such as horses, bikes and cars.  We attribute this diversity of categories from the discriminator as a core reason why the encoder of our \textit{UFO Search} method is able to learn to retrieve compatible objects from multiple categories, as shown in the \textit{UFO} retrieval of the last example.  The discriminator can effectively generate triplets for training the encoder.  In contrast, \textit{No Discriminator} only retrieves cars although multiple object types are appropriate for the scene.  Further analysis of the retrieval diversity is in the Supplementary Materials.

\section{Conclusion}
\vspace{-0.2em}
We introduce a novel problem of searching for compatible foreground objects to edit images without constraints on object types.  We also propose a solution with an efficient, scalable approach for generating a large training dataset.  Experiments demonstrate advantages of our approach for efficiently and accurately retrieving compatible foreground objects from large-scale, diverse datasets.  We offer this work to support people in efficiently editing their images.

\vspace{-1em}
\paragraph{Acknowledgments.} We thank the anonymous reviewers for their feedback and crowd workers for their contributions.  This work is supported in part by gifts from Adobe.

\clearpage

\balance{}
{\small
\bibliographystyle{ieee_fullname}
\bibliography{egbib}

\begin{thebibliography}{10}\itemsep=-1pt

\bibitem{ak2018learning}
Kenan~E Ak, Ashraf~A Kassim, Joo Hwee~Lim, and Jo Yew~Tham.
\newblock Learning attribute representations with localization for flexible
  fashion search.
\newblock In {\em Proceedings of the IEEE Conference on Computer Vision and
  Pattern Recognition}, pages 7708--7717, 2018.

\bibitem{azadi2018compositional}
Samaneh Azadi, Deepak Pathak, Sayna Ebrahimi, and Trevor Darrell.
\newblock Compositional gan: Learning conditional image composition.
\newblock {\em arXiv preprint arXiv:1807.07560}, 2018.

\bibitem{bar1996spatial}
Moshe Bar and Shimon Ullman.
\newblock Spatial context in recognition.
\newblock {\em Perception}, 25(3):343--352, 1996.

\bibitem{Barnes:2009:PAR}
Connelly Barnes, Eli Shechtman, Adam Finkelstein, and Dan~B Goldman.
\newblock {PatchMatch}: A randomized correspondence algorithm for structural
  image editing.
\newblock {\em ACM Transactions on Graphics (Proc. SIGGRAPH)}, 28(3), Aug.
  2009.

\bibitem{bell2016inside}
Sean Bell, C Lawrence~Zitnick, Kavita Bala, and Ross Girshick.
\newblock Inside-outside net: Detecting objects in context with skip pooling
  and recurrent neural networks.
\newblock In {\em Proceedings of the IEEE conference on computer vision and
  pattern recognition}, pages 2874--2883, 2016.

\bibitem{divvala2009empirical}
Santosh~K Divvala, Derek Hoiem, James~H Hays, Alexei~A Efros, and Martial
  Hebert.
\newblock An empirical study of context in object detection.
\newblock In {\em Computer Vision and Pattern Recognition, 2009. CVPR 2009.
  IEEE Conference on}, pages 1271--1278. IEEE, 2009.

\bibitem{hays2007scene}
James Hays and Alexei~A Efros.
\newblock Scene completion using millions of photographs.
\newblock In {\em ACM Transactions on Graphics (TOG)}, volume~26, page~4. ACM,
  2007.

\bibitem{iizuka2017globally}
Satoshi Iizuka, Edgar Simo-Serra, and Hiroshi Ishikawa.
\newblock Globally and locally consistent image completion.
\newblock {\em ACM Transactions on Graphics (ToG)}, 36(4):107, 2017.

\bibitem{JDH17}
Jeff Johnson, Matthijs Douze, and Herv{\'e} J{\'e}gou.
\newblock Billion-scale similarity search with gpus.
\newblock {\em arXiv preprint arXiv:1702.08734}, 2017.

\bibitem{kingma2014adam}
Diederik~P Kingma and Jimmy Ba.
\newblock Adam: A method for stochastic optimization.
\newblock {\em arXiv preprint arXiv:1412.6980}, 2014.

\bibitem{lalonde2007photo}
Jean-Fran{\c{c}}ois Lalonde, Derek Hoiem, Alexei~A Efros, Carsten Rother, John
  Winn, and Antonio Criminisi.
\newblock Photo clip art.
\newblock {\em ACM transactions on graphics (TOG)}, 26(3):3, 2007.

\bibitem{lin2018st}
Chen-Hsuan Lin, Ersin Yumer, Oliver Wang, Eli Shechtman, and Simon Lucey.
\newblock St-gan: Spatial transformer generative adversarial networks for image
  compositing.

\bibitem{lin2014microsoft}
Tsung-Yi Lin, Michael Maire, Serge Belongie, James Hays, Pietro Perona, Deva
  Ramanan, Piotr Doll{\'a}r, and C~Lawrence Zitnick.
\newblock Microsoft coco: Common objects in context.
\newblock In {\em European conference on computer vision}, pages 740--755.
  Springer, 2014.

\bibitem{paszke2017automatic}
Adam Paszke, Sam Gross, Soumith Chintala, Gregory Chanan, Edward Yang, Zachary
  DeVito, Zeming Lin, Alban Desmaison, Luca Antiga, and Adam Lerer.
\newblock Automatic differentiation in pytorch.
\newblock 2017.

\bibitem{pathak2016context}
Deepak Pathak, Philipp Krahenbuhl, Jeff Donahue, Trevor Darrell, and Alexei~A
  Efros.
\newblock Context encoders: Feature learning by inpainting.
\newblock In {\em Proceedings of the IEEE Conference on Computer Vision and
  Pattern Recognition}, pages 2536--2544, 2016.

\bibitem{perez2003poisson}
Patrick P{\'e}rez, Michel Gangnet, and Andrew Blake.
\newblock Poisson image editing.
\newblock {\em ACM Transactions on graphics (TOG)}, 22(3):313--318, 2003.

\bibitem{rodriguez2016data}
Jose~A Rodriguez-Serrano, Diane Larlus, and Zhenwen Dai.
\newblock Data-driven detection of prominent objects.
\newblock {\em IEEE transactions on pattern analysis and machine intelligence},
  38(10):1969--1982, 2016.

\bibitem{russakovsky2015imagenet}
Olga Russakovsky, Jia Deng, Hao Su, Jonathan Krause, Sanjeev Satheesh, Sean Ma,
  Zhiheng Huang, Andrej Karpathy, Aditya Khosla, Michael Bernstein, et~al.
\newblock Imagenet large scale visual recognition challenge.
\newblock {\em International Journal of Computer Vision}, 115(3):211--252,
  2015.

\bibitem{simonyan2014very}
Karen Simonyan and Andrew Zisserman.
\newblock Very deep convolutional networks for large-scale image recognition.
\newblock {\em arXiv preprint arXiv:1409.1556}, 2014.

\bibitem{strat1991context}
Thomas~M Strat and Martin~A Fischler.
\newblock Context-based vision: recognizing objects using information from both
  2 d and 3 d imagery.
\newblock {\em IEEE Transactions on Pattern Analysis and Machine Intelligence},
  13(10):1050--1065, 1991.

\bibitem{sun2017seeing}
Jin Sun and David~W Jacobs.
\newblock Seeing what is not there: Learning context to determine where objects
  are missing.
\newblock In {\em 2017 IEEE Conference on Computer Vision and Pattern
  Recognition (CVPR)}, pages 1234--1242. IEEE, 2017.

\bibitem{sunkavalli2010multi}
Kalyan Sunkavalli, Micah~K Johnson, Wojciech Matusik, and Hanspeter Pfister.
\newblock Multi-scale image harmonization.
\newblock In {\em ACM Transactions on Graphics (TOG)}, volume~29, page 125.
  ACM, 2010.

\bibitem{tan2018and}
Fuwen Tan, Crispin Bernier, Benjamin Cohen, Vicente Ordonez, and Connelly
  Barnes.
\newblock Where and who? automatic semantic-aware person composition.
\newblock In {\em 2018 IEEE Winter Conference on Applications of Computer
  Vision (WACV)}, pages 1519--1528. IEEE, 2018.

\bibitem{tao2010error}
Michael~W Tao, Micah~K Johnson, and Sylvain Paris.
\newblock Error-tolerant image compositing.
\newblock In {\em European Conference on Computer Vision}, pages 31--44.
  Springer, 2010.

\bibitem{torralba2003contextual}
Antonio Torralba.
\newblock Contextual priming for object detection.
\newblock {\em International journal of computer vision}, 53(2):169--191, 2003.

\bibitem{torralba200880}
Antonio Torralba, Rob Fergus, and William~T Freeman.
\newblock 80 million tiny images: A large data set for nonparametric object and
  scene recognition.
\newblock {\em IEEE transactions on pattern analysis and machine intelligence},
  30(11):1958--1970, 2008.

\bibitem{tsai2017deep}
Yi-Hsuan Tsai, Xiaohui Shen, Zhe Lin, Kalyan Sunkavalli, Xin Lu, and Ming-Hsuan
  Yang.
\newblock Deep image harmonization.
\newblock In {\em IEEE Conference on Computer Vision and Pattern Recognition
  (CVPR)}, volume~2, 2017.

\bibitem{wang2015unsupervised}
Xiaolong Wang and Abhinav Gupta.
\newblock Unsupervised learning of visual representations using videos.
\newblock In {\em Proceedings of the IEEE International Conference on Computer
  Vision}, pages 2794--2802, 2015.

\bibitem{IntRetrieval2009}
Oliver Whyte, Josef Sivic, and Andrew Zisserman.
\newblock Get out of my picture! internet-based inpainting.
\newblock In {\em BMVC}, volume~2, page~5, 2009.

\bibitem{yang2016high}
Chao Yang, Xin Lu, Zhe Lin, Eli Shechtman, Oliver Wang, and Hao Li.
\newblock High-resolution image inpainting using multi-scale neural patch
  synthesis.
\newblock In {\em The IEEE Conference on Computer Vision and Pattern
  Recognition (CVPR)}, volume~1, page~3, 2017.

\bibitem{yang2014context}
Jimei Yang, Brian Price, Scott Cohen, and Ming-Hsuan Yang.
\newblock Context driven scene parsing with attention to rare classes.
\newblock In {\em Proceedings of the IEEE conference on computer vision and
  pattern recognition}, pages 3294--3301, 2014.

\bibitem{yu2018generative}
Jiahui Yu, Zhe Lin, Jimei Yang, Xiaohui Shen, Xin Lu, and Thomas~S Huang.
\newblock Generative image inpainting with contextual attention.
\newblock {\em arXiv preprint}, 2018.

\bibitem{zhao2018compositing}
Hengshuang Zhao, Xiaohui Shen, Zhe Lin, Kalyan Sunkavalli, Brian Price, and
  Jiaya Jia.
\newblock Compositing-aware image search.
\newblock In {\em Proceedings of the European Conference on Computer Vision
  (ECCV)}, pages 502--516, 2018.

\bibitem{zhao2018guided}
Yinan Zhao, Brian Price, Scott Cohen, and Danna Gurari.
\newblock Guided image inpainting: Replacing an image region by pulling content
  from another image.
\newblock {\em arXiv preprint arXiv:1803.08435}, 2018.

\bibitem{zhu2015learning}
Jun-Yan Zhu, Philipp Krahenbuhl, Eli Shechtman, and Alexei~A Efros.
\newblock Learning a discriminative model for the perception of realism in
  composite images.
\newblock In {\em Proceedings of the IEEE International Conference on Computer
  Vision}, pages 3943--3951, 2015.

\bibitem{zhu2015faithful}
Zhe Zhu, Hao-Zhi Huang, Zhi-Peng Tan, Kun Xu, and Shi-Min Hu.
\newblock Faithful completion of images of scenic landmarks using internet
  images.
\newblock {\em IEEE transactions on visualization and computer graphics},
  22(8):1945--1958, 2015.

\end{thebibliography}
}

\clearpage
\onecolumn
\newpage
\appendix
\noindent {\LARGE \textbf{Appendix}}
\vspace{1em}

This document supplements our methods and results provided in the main paper. In Section~\ref{sampling}, we describe the sampling module used to accelerate identifying plausible foreground objects for training the encoder (supplements Section \textbf{3.3}). In Section~\ref{decomposition}, we describe how we decompose an image into a background scene and foreground object when we create the dataset from MS-COCO~\cite{lin2014microsoft} (supplements Section \textbf{4.2}).  In Section~\ref{sec:cais}, we show more quantitative and qualitative results in CAIS~\cite{zhao2018compositing} (supplements Section \textbf{4.1}). In Section~\ref{sec:mscoco}, we show more qualitative results in MS-COCO~\cite{lin2014microsoft} (supplements Section \textbf{4.2}). In Section~\ref{sec:diversity}, we show a quantitative analysis of retrieval diversity in MS-COCO~\cite{lin2014microsoft} (supplements Section \textbf{4.2}). In Section~\ref{sec:post-processing}, we show two applications of our \textit{UFO Search}: hole-filling and compositing (supplements Figure \textbf{1} in the main paper).

\section{Sampling Module (supplements Section \textbf{3.3} in main paper)}
\label{sampling}
To speed up the discriminator's role in generating training data, in the main paper we introduce a sampling module to accelerate sampling plausible foreground objects for training the encoder. Specifically, we put in the sampling set some random foreground objects, the top $K_C$ most similar background scenes and top $K_G$ most similar foreground objects. To measure the similarity between objects and background scenes, we use VGG-19~\cite{simonyan2014very} pretrained for the ILSVRC-2014 competition~\cite{russakovsky2015imagenet} to extract features for foreground objects and background scenes respectively. We normalize the feature vectors to unit length and measure the similarities between objects and between background scenes by cosine similarity. To speed up similarity search, we build an index for the foreground object database and background scene database separately by Faiss~\cite{JDH17}, a library for efficient similarity search and clustering of dense vectors.

\section{Decomposition of Foreground and Background (supplements Section \textbf{4.2} in main paper)}
\label{decomposition}
 In Section \textbf{4.2} of the main paper, we employ MS-COCO~\cite{lin2014microsoft} to create a diverse dataset of 79 foreground object categories for evaluation. We use the annotated object instance segmentation mask to decompose each image into foreground objects and a background scene. Specifically, to create a foreground image of input size 224x224, we segment out the foreground object using the instance segmentation mask, and then overlay it at the center of a square 224x224 image consisting of pixels set to the (ImageNet) image mean.  As for the background image, we first crop it to the size 224x224 that is required by our network.  We then create the hole by removing the original object in the annotated object instance segmentation mask as well as all pixels in a bounding rectangle around the object. We then set the pixels in the hole to the (ImageNet) image mean.

\section{Quantitative and Qualitative Results in CAIS (supplements Section \textbf{4.1} in main paper)}
\label{sec:cais}

In Section \textbf{4.1} of the main paper, we evaluate our method quantitatively in CAIS~\cite{lin2014microsoft}. In this section, we show more quantitative and qualitative results in CAIS~\cite{lin2014microsoft}. 

In Table~\ref{table:MAP}, we show mAP results  with respect to all the retrievals for two baselines, \textit{UFO Search} and ablated variants of \textit{UFO Search} in CAIS~\cite{zhao2018compositing}. Note that we do not show results of \textit{CFO} methods because they do not rank all objects in all categories. Overall, our \textit{UFO Search} outperforms other baselines and variants; e.g. mAP is \textbf{29.03} for \textit{UFO Search}, which is \textbf{3.34} percentage points improvement over the next best ablated variant. \textit{UFO Search} outperforms other baselines and ablated variants in the following five object categories: boat, bottle, car, painting and plant. These findings align with and reinforce those in the main paper. 

\begin{figure*}[!ht]
\centering
\includegraphics[width=0.95\textwidth]{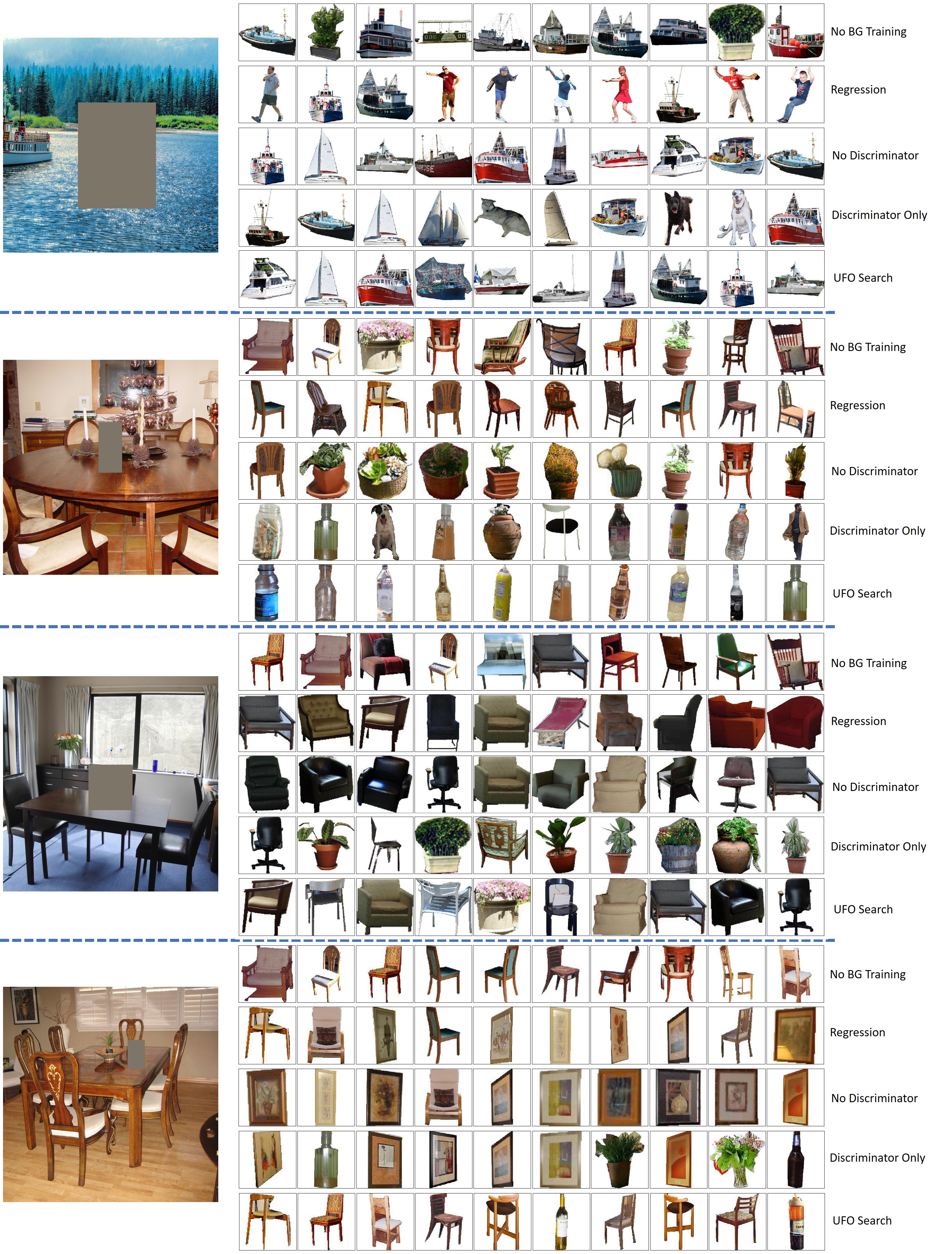}
\caption{Qualitative results in CAIS~\cite{zhao2018compositing}. In the top two examples, \textit{UFO Search} retrieves objects from the only compatible object type in the eight categories (boat and bottle, respectively). The last two examples demonstrate failure cases where our method fails to capture the spatial relationship between inserted objects and existing objects in the scene.}
\label{fig:cais_results}
\end{figure*}

We also show qualitative results in Figure \ref{fig:cais_results}. In the top two examples, our \textit{UFO Search} retrieves objects from the only compatible object type in the eight categories of CAIS~\cite{lin2014microsoft} (boat and bottle, respectively). Note that the compatible foreground objects depend on both hole shape and context in this dataset with various hole shapes. The bottom two examples demonstrate failure cases. For the second last example, our method retrieves chairs for a hole where there is supposed to be a plant. For the last example, our approach also retrieves chairs to fill the hole where bottles are compatible. Chairs are reasonable objects to be present in both scenes. However, they are not compatible in the given holes that are placed on top of tables. In other words, at times, our method fails to capture the spatial relationship between inserted objects and existing objects in both scenes. Our current model does not explicitly model spatial relationship between objects. It is interesting to explore in future work how to model the spatial relationship between objects more effectively.

\begin{table*}[!ht]
\label{table:MAP}
\begin{center}
\begin{tabular}{l | c c c c c c c c | c}
\hline
Method & boat & bottle & car & chair & dog & painting & person & plant & overall \\
\hline\hline
Shape~\cite{zhao2018compositing} & 6.53 & 3.69 & 9.90 & 4.09 & 10.89 & 2.56 & 4.74 & 5.55 & 5.99 \\
RealismCNN~\cite{zhu2015learning} & 5.84 & 6.10 & 3.55 & 1.50 & 5.68 & 2.66 & 3.27 & 7.58 & 4.52 \\
Ours: No BG Training  & 34.19 & 4.61 & 3.98 & 4.84 & 10.69 & 8.70 & 8.81 & 16.48 & 11.54\\ 
Ours: Discriminator Only & 32.99 & 8.68 & 17.67 & 13.13 & \textbf{35.67} & 16.29 & 17.58 & 13.62 & 19.45\\
Ours: Regression & 49.31 & 11.82 & 16.98 & 12.09 & 28.07 & 21.35 & \textbf{30.11} & 10.44 & 22.52\\
Ours: No Discriminator & 52.85 & 16.94 & 19.62 & \textbf{16.13} & 31.92 & 21.17 & 24.11 & 22.80 & 25.69\\
Ours: UFO Search & \textbf{56.64} & \textbf{23.62} & \textbf{31.63} & 13.77 & 33.39 & \textbf{24.33} & 23.94 & \textbf{24.93} & \textbf{29.03}\\
\hline
\end{tabular}
\end{center}
\caption{\textit{Mean Average Precision} with respect to all the retrievals.  Results are shown as percentages.
}
\label{table:MAP}
\end{table*}

\section{Qualitative Results in MS-COCO (supplements Section \textbf{4.2} in main paper)}
\label{sec:mscoco}

\begin{figure*}[!ht]
\centering
\includegraphics[width=0.93\textwidth]{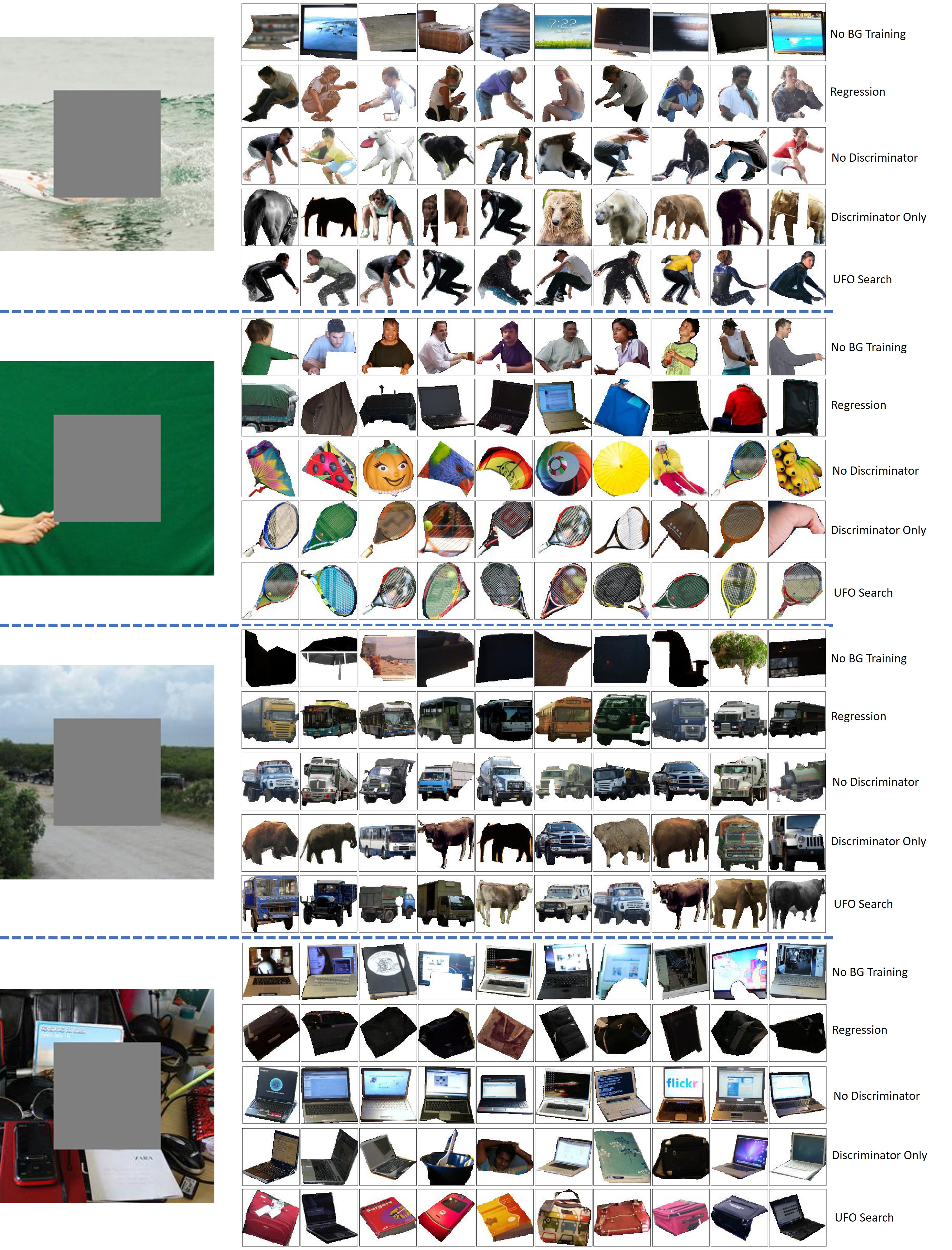}
\caption{Qualitative results in MS-COCO~\cite{lin2014microsoft}. In the first two examples, \textit{UFO Search} retrieves objects from the only object type in MS-COCO (surfer and racket, respectively) that is really compatible with the context. The bottom two examples show that our method can retrieve compatible objects from differing categories when numerous object types are appropriate for the scene. For example, our method retrieves truck, jeep and cattle for a hole on a road (second to bottom example) and retrieves bag, book and laptop for a hole on a cluttered table (bottom example).}
\label{fig:coco_results1}
\end{figure*}

\begin{figure*}[!ht]
\centering
\includegraphics[width=0.95\textwidth]{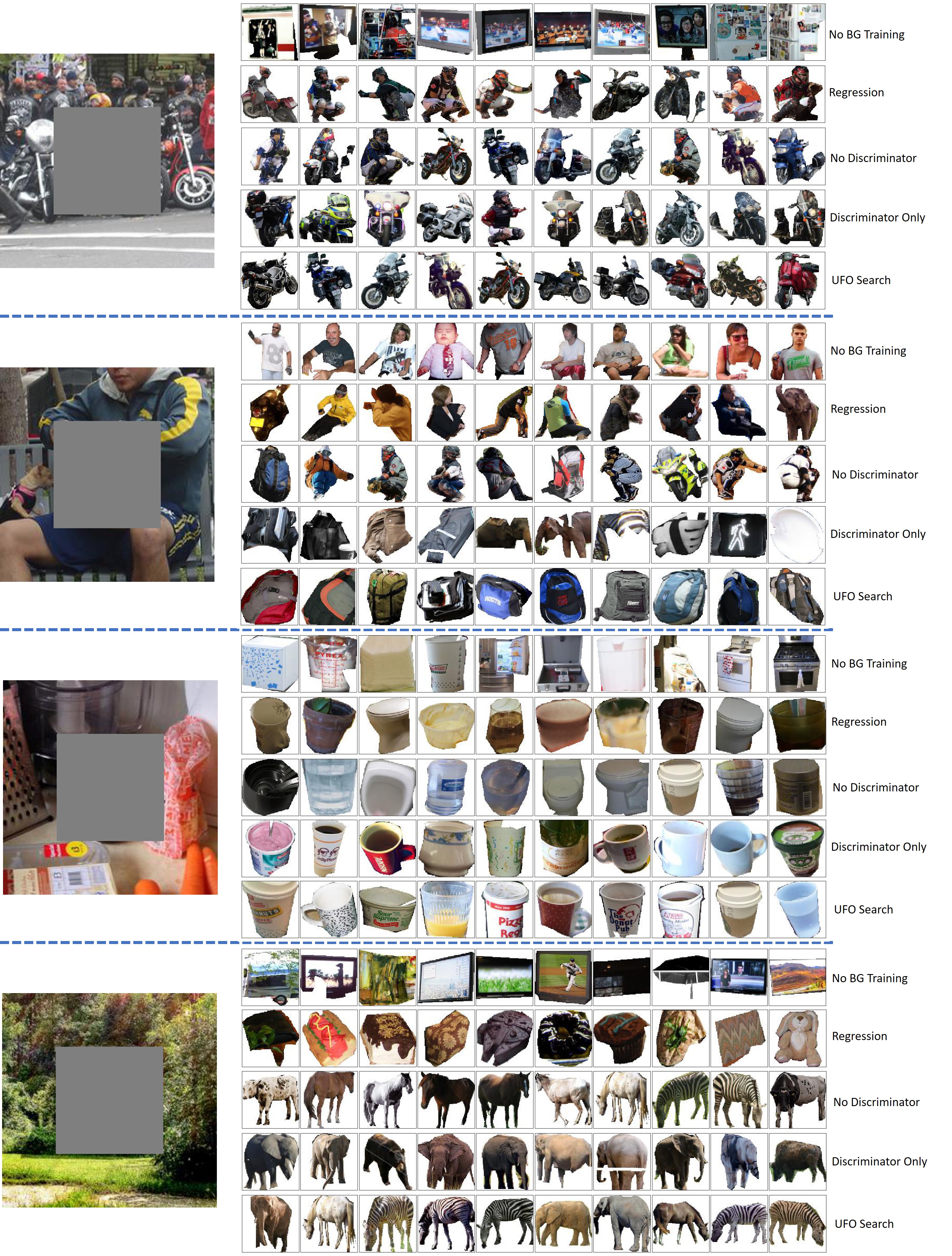}
\caption{Qualitative results in MS-COCO~\cite{lin2014microsoft}. In the top three examples, \textit{UFO Search} retrieves objects from the best fitting category in MS-COCO (motorcycle, bag and cup, respectively) in the top-ranked retrievals. In the bottom example, \textit{UFO Search} retrieves compatible objects from differing categories, e.g. horse, zebra and elephant.}
\label{fig:coco_results2}
\end{figure*}

In the main paper, we conduct user evaluation and show qualitative results in MS-COCO~\cite{lin2014microsoft}. We show more qualitative results in Figure \ref{fig:coco_results1} and Figure \ref{fig:coco_results2} to exemplify our performance in the more diverse dataset consisting of 79 categories. If there is only one object type that is compatible with the context, our method retrieves only those objects. Our approach also has the potential to retrieve compatible objects of different categories when many object types are appropriate for the scene.

\section{Quantitative Analysis of Retrieval Diversity (supplements Section \textbf{4.2} in main paper)}
\label{sec:diversity}
In Figure \textbf{5} of the main paper, we qualitatively demonstrate that our \textit{UFO Search} has the potential to retrieve compatible objects of different categories when many object types are appropriate for the scene. In this section, we measure retrieval diversity quantitatively.  To do so, we compute the average number of categories compatible objects span in the top 25 retrievals of our user study in Section \textbf{4.2}. Results are shown in Table~\ref{table:diversity}.  Note that we measure diversity on compatible objects only instead of all the retrievals. Otherwise, a random guess would have a large diversity although most of its retrievals would be incompatible.

\begin{table}[h!]
\label{table:coco}
\begin{center}
\begin{tabular}{l|c|c}
\hline
Method & Diversity ($\pm$ Std Dev) & P@25 \\
\hline
 UFO Search & 1.90 ($\pm$ 1.54)  & 38.83 \\
 No Discriminator & 1.82 ($\pm$ 1.30) & 35.57 \\
 Discriminator Only & 2.72 ($\pm$ 3.06) & 35.77 \\
 Regression & 2.56 ($\pm$ 2.34) & 30.40 \\
 No BG Training & 1.52 ($\pm$ 0.79)  & 12.50 \\
 \hline
\end{tabular}
\end{center}
\caption{Quantitative analysis of diversity for our \textit{UFO Search} method and its four ablated variants. We report the average number of categories compatible objects span in the top 25 retrievals along with the standard deviation. For completeness, we also report retrieval performance results (P@25) from Table \textbf{2} in the main paper.
}
\label{table:diversity}
\end{table}

The findings reveal that the increased diversity of \textit{UFO Search} over \textit{No Discriminator} may come from the discriminator, as evidenced by the fact that \textit{Discriminator Only} has the largest diversity among the shown five methods. While \textit{Regression} has the largest diversity in encoder-based methods (\textit{UFO Search}, \textit{No Discriminator}, \textit{Regression} and \textit{No BG Training}), the percentage of compatible objects it retrieves is significantly lower than \textit{UFO Search} and \textit{No Discriminator}.

\section{Applications (supplements Figure 1 in main paper)}
\label{sec:post-processing}
\begin{figure*}[!ht]
\centering
\includegraphics[width=0.93\textwidth]{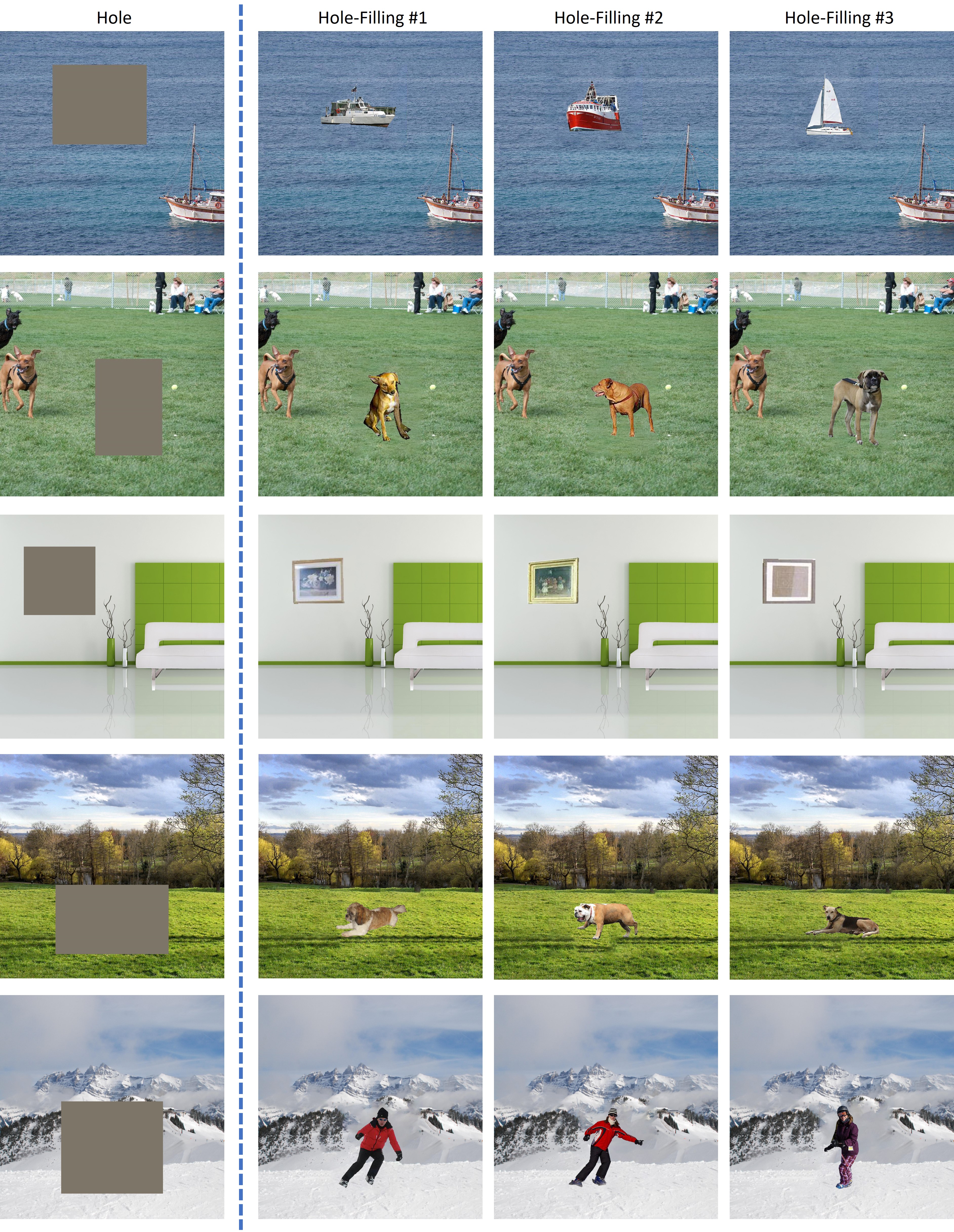}
\caption{Application of \textit{UFO Search} in hole-filling. We show the background image with a hole on the left, and three different hole-filling results on its right. To fill the hole, we pick a top-ranked foreground object retrieved by our \textit{UFO Search}, overlay the object in the center of the hole, and fill the smaller gaps around it using PatchMatch~\cite{Barnes:2009:PAR}. The shown background image and foreground objects are in the test set of CAIS~\cite{zhao2018compositing}. Note that there are not a diversity of object types being retrieved for the shown background images since most background images in CAIS~\cite{zhao2018compositing} unambiguously match only one assigned foreground object category from the few candidate categories represented. }
\label{fig:post}
\end{figure*}

\begin{figure*}[!ht]
\centering
\includegraphics[width=0.92\textwidth]{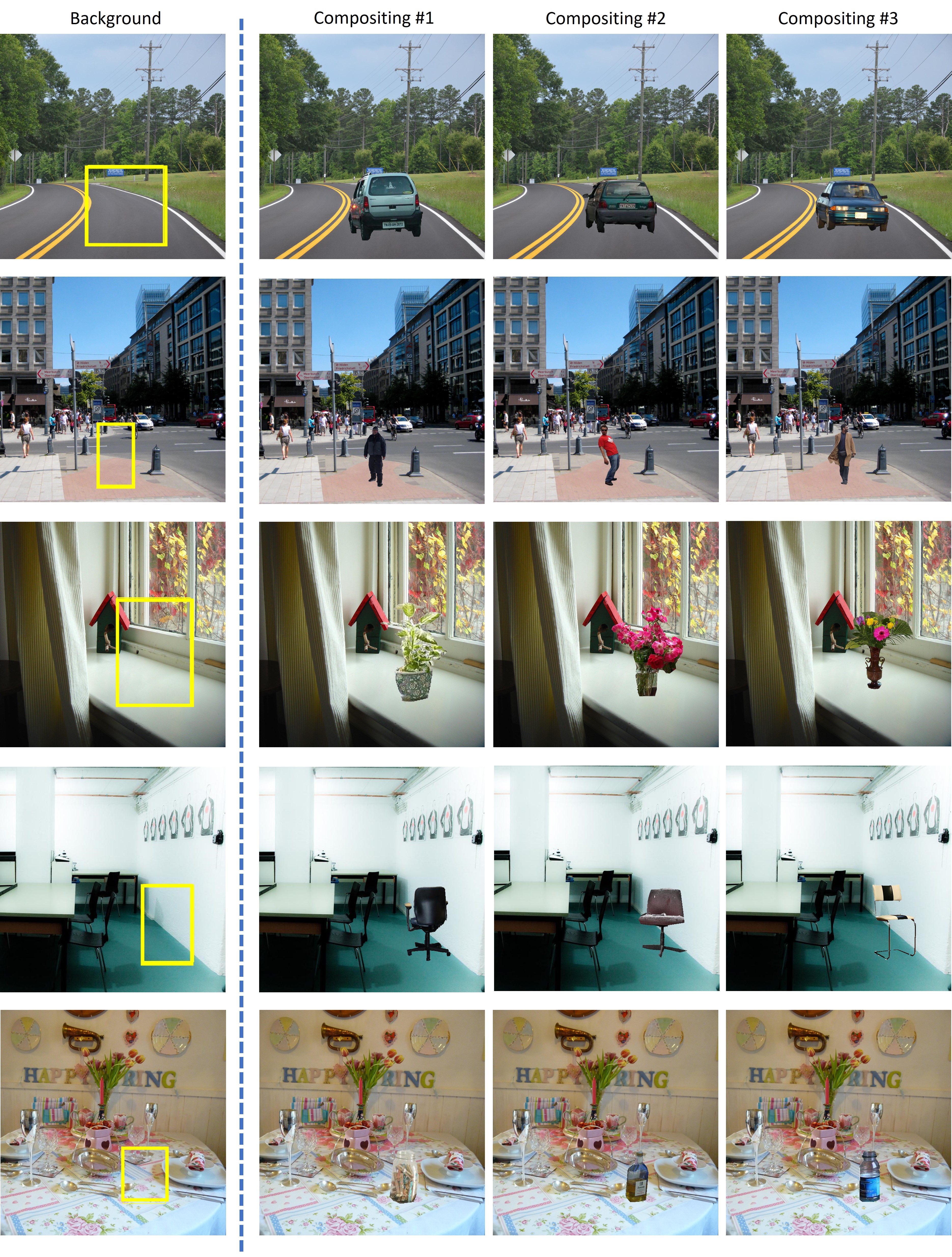}
\caption{Application of \textit{UFO Search} in compositing. We show on the left the background image with a yellow rectangle indicating the position to insert the object, and three different compositing results on its right. The shown background image and foreground objects are in the test set of CAIS~\cite{zhao2018compositing}. Note that there are not a diversity of object types being retrieved for the shown background images since most background images in CAIS~\cite{zhao2018compositing} unambiguously match only one assigned foreground object category from the few candidate categories represented. }
\label{fig:comp}
\end{figure*}

For completeness, we show two applications of \textit{UFO Search}: hole-filling in Figure~\ref{fig:post} and compositing in Figure~\ref{fig:comp}. Specifically, we use our \textit{UFO Search} to retrieve compatible foreground objects, and then use them to aid (1) hole-filling and (2) compositing them directly into the background. For hole-filling, we overlay the object in the center of the hole, and fill the smaller gap around it using PatchMatch~\cite{Barnes:2009:PAR}, as shown in Figure~\ref{fig:post}.  For compositing, we insert the retrieved object at the center of the specified yellow rectangle, as shown in Figure~\ref{fig:comp}. Note that the hole-filling and compositing results would look more natural with harmonization~\cite{sunkavalli2010multi, tao2010error, perez2003poisson, tsai2017deep, zhao2018guided}. We are not applying such methods since that is not the main focus of this paper.

\end{document}